\documentclass[article,twocolumn,showpacs,aps,prl,floatfix,superscriptaddress]{revtex4-1}
\usepackage{times}
\usepackage{color}
\usepackage[colorlinks=true,urlcolor=blue,citecolor=blue]{hyperref}
\usepackage{url}
\usepackage{breakurl}
\usepackage{soul}
\usepackage{graphicx}
\usepackage{amsmath}
\usepackage{subfigure}
\usepackage{xcolor}
\usepackage{amssymb}
\usepackage{latexsym}
\usepackage{hyperref}
\DeclareGraphicsExtensions{.png,.eps}
\usepackage{float}
\usepackage{appendix}
\usepackage{etoolbox}

\usepackage{xcolor}
\usepackage[urlcolor=blue]{hyperref}      
\hypersetup{
	colorlinks = true,                    
	citecolor = {blue},
	linkcolor = {purple},
}

\begin{document}	
	
\title{Deep Machine Learning Reconstructing Lattice Topology with Strong Thermal Fluctuations}

\author{Xiao-Han Wang}
\affiliation{Department of Physics, Capital Normal University, Beijing 100048, China}
\author{Pei Shi}
\affiliation{Department of Physics, Capital Normal University, Beijing 100048, China}
\author{Bin Xi}
\affiliation{College of Physics Science and Technology, Yangzhou University, Yangzhou 225002, China}
\author{Jie Hu}\email[Corresponding author. Email: ]{jie.hu@cnu.edu.cn}
\affiliation{Department of Physics, Capital Normal University, Beijing 100048, China}
\author{Shi-Ju Ran}\email[Corresponding author. Email: ]{sjran@cnu.edu.cn}
\affiliation{Department of Physics, Capital Normal University, Beijing 100048, China}

\date{\today}
\begin{abstract}
Applying artificial intelligence to scientific problems (namely AI for science) is currently under hot debate. However, the scientific problems differ much from the conventional ones with images, texts, and etc., where new challenges emerges with the unbalanced scientific data and complicated effects from the physical setups. In this work, we demonstrate the validity of the deep convolutional neural network (CNN) on reconstructing the lattice topology (i.e., spin connectivities) in the presence of strong thermal fluctuations and unbalanced data. Taking the kinetic Ising model with Glauber dynamics as an example, the CNN maps the time-dependent local magnetic momenta (a single-node feature) evolved from a specific initial configuration (dubbed as an evolution instance) to the probabilities of the presences of the possible couplings. Our scheme distinguishes from the previous ones that might require the knowledge on the node dynamics, the responses from perturbations, or the evaluations of statistic quantities such as correlations or transfer entropy from many evolution instances. The fine tuning avoids the ``barren plateau'' caused by the strong thermal fluctuations at high temperatures. Accurate reconstructions can be made where the thermal fluctuations dominate over the correlations and consequently the statistic methods in general fail. Meanwhile, we unveil the generalization of CNN on dealing with the instances evolved from the unlearnt initial spin configurations and those with the unlearnt lattices. We raise an open question on the learning with unbalanced data in the nearly ``double-exponentially'' large sample space. 
\end{abstract}
\maketitle

Booming progresses have been made recently in applying artificial intelligence (AI) to science~\cite{butler_machine_2018,dilsizian_machine_2018,dunjko_machine_2018,carleo_machine_2019}, which permits to solve various problems that cannot be validly solved via the conventional methods. For instance, we usually consider to simulate the physical properties of a system with known parameters (e.g., interactions, temperature, etc.). Machine learning (ML) opens new data-driven avenues to solving the inverse problems~\cite{arel_deep_2010}, such as the predictions of crystal structures~\cite{cubuk_identifying_2015, berthusen_learning_2021} and Hamiltonians~\cite{hegde_machine-learned_2017, hong_predicting_2021, ma_deep_2021,Wang2021} given the relevant data. 

Focusing on the Ising model described by the Glauber dynamics~\cite{glauber_timedependent_1963}, it has served as a fundamental mathematic model in statistic physics~\cite{binder_phase_1976, godreche_response_2000, bukharov_magnetic_2009, kaneyoshi_magnetizations_2009}, sociology~\cite{drouffe_phase_1999, balankin_ising_2017}, biology~\cite{baake_ising_1997, de_oliveira_dynamic_2013}, and etc. A common practice with the Ising model (and many others) for investigating the interested phenomena is to simulate with priorly assumed parameters. Considering that the dynamic or thermodynamic data are accessible in many realistic or experimental scenarios, efficient methods for estimating the parameters of the model (particularly the connectivities among the Ising spins) from the observed data are strongly desired.

Massive efforts have been made in reconstructing the lattice topology from data (see, for instance, Refs.~[\onlinecite{PhysRevLett.97.188701, PhysRevLett.98.224101, Shandilya_2011, Pert2016, nitzan_revealing_2017, peixoto_network_2019,schreiber_measuring_2000, lau_information_2013, barnett_information_2013, deng_renyi_2014, casadiego_model-free_2017, goetze_reconstructing_2019, zeng_network_2011, ching_reconstructing_2015, dettmer_network_2016, ching_reconstructing_2017, lai_reconstructing_2017, donner_inverse_2017, tam_reconstructing_2018}]). The previous methods require, e.g., certain priori knowledge on the node dynamics, the responses by implementing the designed perturbations, or the sparsity of the connections~\cite{PhysRevLett.97.188701, PhysRevLett.98.224101, Shandilya_2011, Pert2016, nitzan_revealing_2017, peixoto_network_2019}. Such requirements limit their applications in the realistic scenarios. A different but related kind of approaches are developed to reconstruct the topology by estimating the statistic quantities from data, such as the correlations among the nodes, likelihood function, or the transfer entropy~\cite{schreiber_measuring_2000, lau_information_2013, barnett_information_2013, deng_renyi_2014, casadiego_model-free_2017, donner_inverse_2017, goetze_reconstructing_2019}. The validity of such methods strongly relies on the relevance between the topology and the statistic quantities~\cite{BLMCH06}, as well as the precision of estimating  such quantities. It remains to be open whether AI, in particular ML, could give birth to robust methods with less reliance on data and priori knowledge. One may refer to some pioneering works in Refs.~[\onlinecite{NIPS2016_3147da8a, pmlr-v80-kipf18a, Zhang2019, chen2020inference, Murphy2021}].

\begin{figure*}[tbp]
	\centering
	\includegraphics[angle=0,width=1\linewidth]{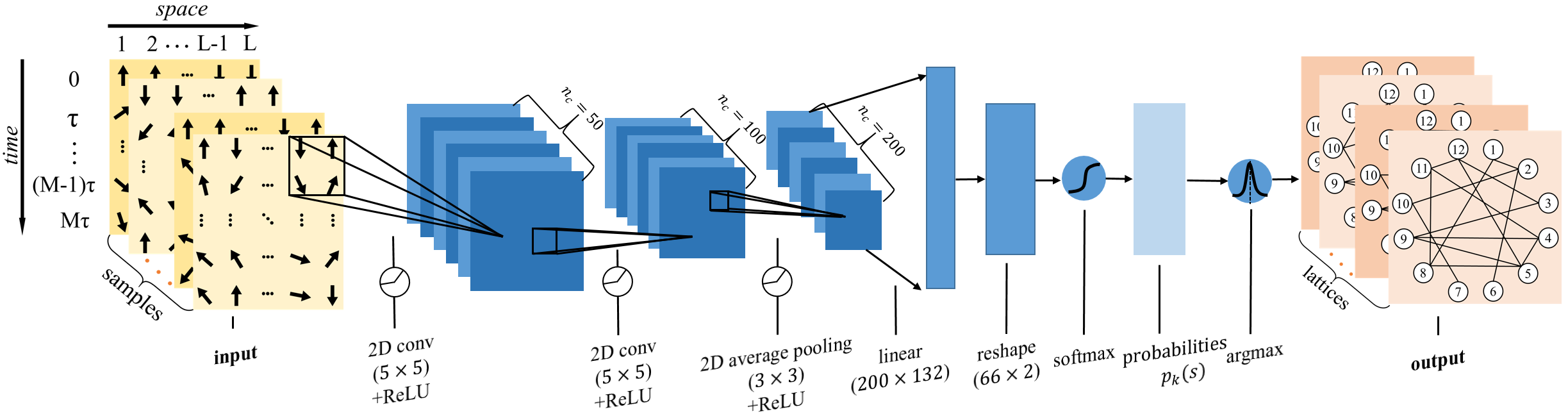}
	\caption{(Color online) The illustration of CNN that maps the magnetic momenta to the probabilities on the presences of possible spins couplings.}
	\label{fig-CNN}
\end{figure*}

In this work, we propose to reconstruct the lattice topology by deep convolutional neural network (CNN). The CNN maps the time-and-space magnetic momenta (a single-node feature) in an evolution from a specific initial state to the probabilities on the presences of the possible Ising couplings. See the illustration of our idea and the structure of the CNN in Fig.~\ref{fig-CNN}. The evolution data from one specific initial state is called an evolution instance. Our scheme requires neither the priori information on the node dynamics nor any responses from perturbations. The trained CNN acts as a ``network generator''~\cite{Zhang2019} and predicts the lattice topology from just one instance. Thus in making the prediction of the connectivities, our method does not require many instances to evaluate the sophisticated statistic quantities such as the correlations~\cite{PhysRevLett.98.224101, Shandilya_2011} or transfer entropy~\cite{lau_information_2013, goetze_reconstructing_2019}.  

At the high temperatures where the thermal fluctuations dominate over the correlations, the perturbative and statistic methods in general become invalid or inaccurate. We show the emergence of a ``barren plateau''~\cite{McClean2018, Wang2021} caused by the strong thermal fluctuations. By fine-tuning the CNN trained with the low-temperature data, we obtain accurate predictions on the topology even at very high temperatures, and the barren plateau is avoided. Finally, we reveal the generalization of CNN on dealing with the instances that are evolved from the unlearnt initial configurations and those with the unlearnt lattices. We argue that the topology reconstruction from single instance is faced with a big challenge on the ``double-exponential'' complexity of the sample space and on learning with unbalanced data.  


\textit{Method.---} Considering the kinetic Ising model that obeys the Glauber dynamics, the magnetic momenta satisfy
\begin{eqnarray}
	  s_{i}[(m+1) \tau]= &&\left[ -s_{i}(m\tau) + \frac {\tanh \frac{T}{2}}{\sum\limits_{j'} A_{ij'}}  \sum\limits_j A_{ij} s_{j}(m\tau) \right]\tau \nonumber \\ && + s_{i}(m\tau),
	\label{eq-1}
\end{eqnarray}
where $s_{i}(m\tau)$ denotes the magnetic momentum of the $i$-th spin at the time $t=m\tau$, $\tau$ characterizes the discretization of time, and $T$ is the temperature. The element of the adjacent matrix $A_{ij}$ indicates the coupling between the $i$-th and $j$-th spins. For simplicity, $A_{ij}=1$ means that the $i$-th and $j$-th spins are connected (coupled), and $A_{ij}=0$ means no connection. We take the Boltzmann constant $k_{B}=1$ for convenience. 

The goal is to predict the connections among the $L$ Ising spins, i.e., the adjacency matrix $A_{ij}$, provided with the space-and-time-dependent magnetic momenta $\{s_{I}(m\tau)\}$ for $i=1, \cdots, L$ and $m=0,\cdots,M-1$. The total evolution time satisfies $t^{\text{tot}}=(M-1)\tau$. The $\{s_{I}(m\tau)\}$ evolved from a specific initial state $s_{i}(0)=\pm{1}$ is dubbed as an evolution instance, which is a ($L \times M$)-dimensional matrix. With the same lattice, different initial state will result in different instances. There are in total $2^L$ possible initial states.

We propose to employ a deep CNN that maps an instance $\{s_{i}(n\tau)\}$ to the adjacent matrix $A_{ij}$. For simplicity, we consider the symmetric (undirected) cases with $A_{ij}=A_{ji}$, and assume the diagonal terms to be $A_{ii}=0$. Therefore, the prediction of the connectivity only concerns the upper-diagonal terms of $A_{ij}$, which in total correspond to $K=\frac{1}{2}L(L-1)$ possible connections. The output of the CNN is $\frac{1}{2}L(L-1)$ probability distributions $\{p_{k}(s)\}$. The values of $p_{k}(s=0)$ and $p_{k}(s=1)$ give the probabilities of having and not having a connection between the $k$-th pair of spins.  The $k$-th connection is predicted as $\arg\max_s[p_{k}(s)]$.

A key advantage of the ML methods is that one does not have to use all possible instances to optimize the model. The ML model possesses certain ability of provide accurate predictions from the unlearnt instances, which is known as the generalization. The training set is here defined as ${N}_{\text{train}}$ instances (or samples in the terminology of ML) used to train the CNN, with ${N}_{\text{train}}<2^L$. The initial states of the training instances are (randomly) taken and are different to each other. We employ the negative-logarithmic likelihood as the loss function, which is defined as
 \begin{eqnarray}
	f = -\frac{2}{NL(L-1)} \sum_{n=1}^{{N}_{\text{train}}} \sum_{k=1}^{\frac{1}{2}L(L-1)} \log p_{k}(s=q_{k}) ,
	\label{eq-3}
\end{eqnarray}
with $q_{k}$ the ground truth of existence of the $k$-th connection. The CNN is optimized by minimizing the loss function following the standard back propagation procedure of neural networks. A trained CNN is expected to give the adjacent matrix of the lattice, requiring only one instance, which makes our scheme more experimentally friendly than the statistic methods such as those based on the transfer entropy.

\textit{Numerical results: confidence, barren plateau, and generalization.---} We use the lattices that generate the training set to generate the testing instances. The initial states of the testing set are randomly taken and are different from those of the training set. In this way, the training and testing sets obey the independent and identical distribution (\textit{i.i.d.}), which is usually assumed to hold in ML. The testing set is used to characterize the accuracy of the CNN on reconstructing the lattice topology from the unlearnt instances. 

\begin{figure}[tbp]
	\centering
	\includegraphics[angle=0,width=0.95\linewidth]{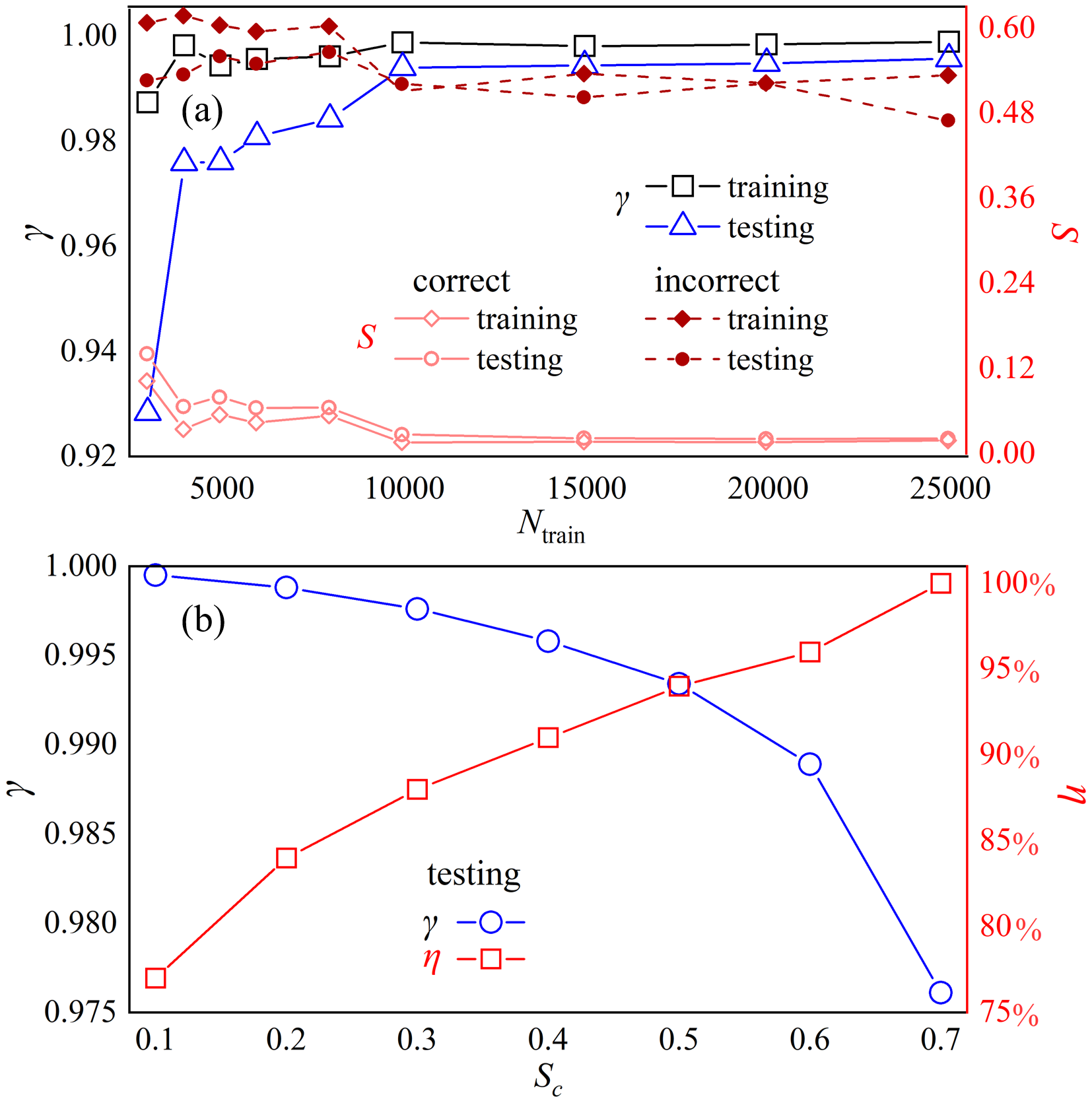}
	\caption{(Color online) (a) The left y-axis shows the training and testing accuracy ($\gamma$) versus the number of training instances ($N_{\text{train}}$) at $T=0.4$. The right y-axis shows the von Neumann entropy $S$ of the probabilities for the correctly and incorrectly predicted connections. (b) For different entropy threshold $S_{c}$, we show the number of connections with $S<S_{c}$ divided by the total number of connections ($\eta$) and the testing accuracy for the connections with $S<S_{c}$.}
	\label{fig-2}
\end{figure}

Fig. \ref{fig-2} (a) shows the accuracy $\gamma$ in the training and testing sets, which is defined as the number of the correct predictions divided by the total number of possible connections. The training and testing set are generated by $N_{L}=10$ different lattices at $T=0.4$. Without losing generality, the connections in each lattice are taken randomly, while the total number of connections in a lattice is restricted to be 25. We take $L=12$ and $T=0.4$, and vary the total number of training instances $N_{\text{train}}$ from $2500$ to $25000$. For each lattice, we equally take  $N_{\text{train}}/10$ instances. We fix the total number of the testing instances to be $N_{\text{test}}=3000$. Both the training and testing accuracies increase with $N_{\text{train}}$, where we have $\gamma\simeq 0.999$ and $0.996$, respectively, for $N_{\text{train}}=25000$.

One advantage of outputting the probability distributions $\{p_{k}(s)\}$ is that we can evaluate the confidence using the von Neumann entropy  defined as $S = - \sum_{s=0,1} p_{k}(s) \ln p_{k}(s)$. We separate the predictions into to the correct and incorrect subgroups . Fig. \ref{fig-2} (a) shows the average $S$ per connection of the ``correct'' or ``incorrect'' subsets with $N_{\text{train}}=10000$ (see the right y axis). We find that for both the training and testing sets, the $\{p_{k}(s)\}$ of the correct predictions in general exhibit much lower $S$ than those of the incorrect predictions. 

The large gap between the $S$ for the correct and incorrect predictions can be used to further increase the accuracy. Fig. \ref{fig-2} (b) shows the $\gamma$ of the test instances with $S<S_c$, i.e., those with relatively low entropy. We observe that for $S_c \simeq 0.1$, the accuracy is close to $1$. The red squares show the proportion $\eta$ defined as the number of connections with $S<S_c$ divided by the total number of connections. For $S_c \simeq 0.6$, we have $\eta \simeq 96\%$, meaning we have confident and correct predictions on about $96\%$ possible connections. Such a relation between accuracy and entropy allow us to use $S$ for marking the unconfident predictions as ``anomalies'' and implement further treatments on them in order to improve the accuracy. For instance, we can use the trained CNN to predict the major part of the possible connections (e.g., $96\%$ with $S_c \simeq 0.6$), and leave the anomalies to more sophisticated or consuming methods, or to the humane experts (active learning).

\begin{figure}[tbp]
	\centering
	\includegraphics[angle=0,width=0.95\linewidth]{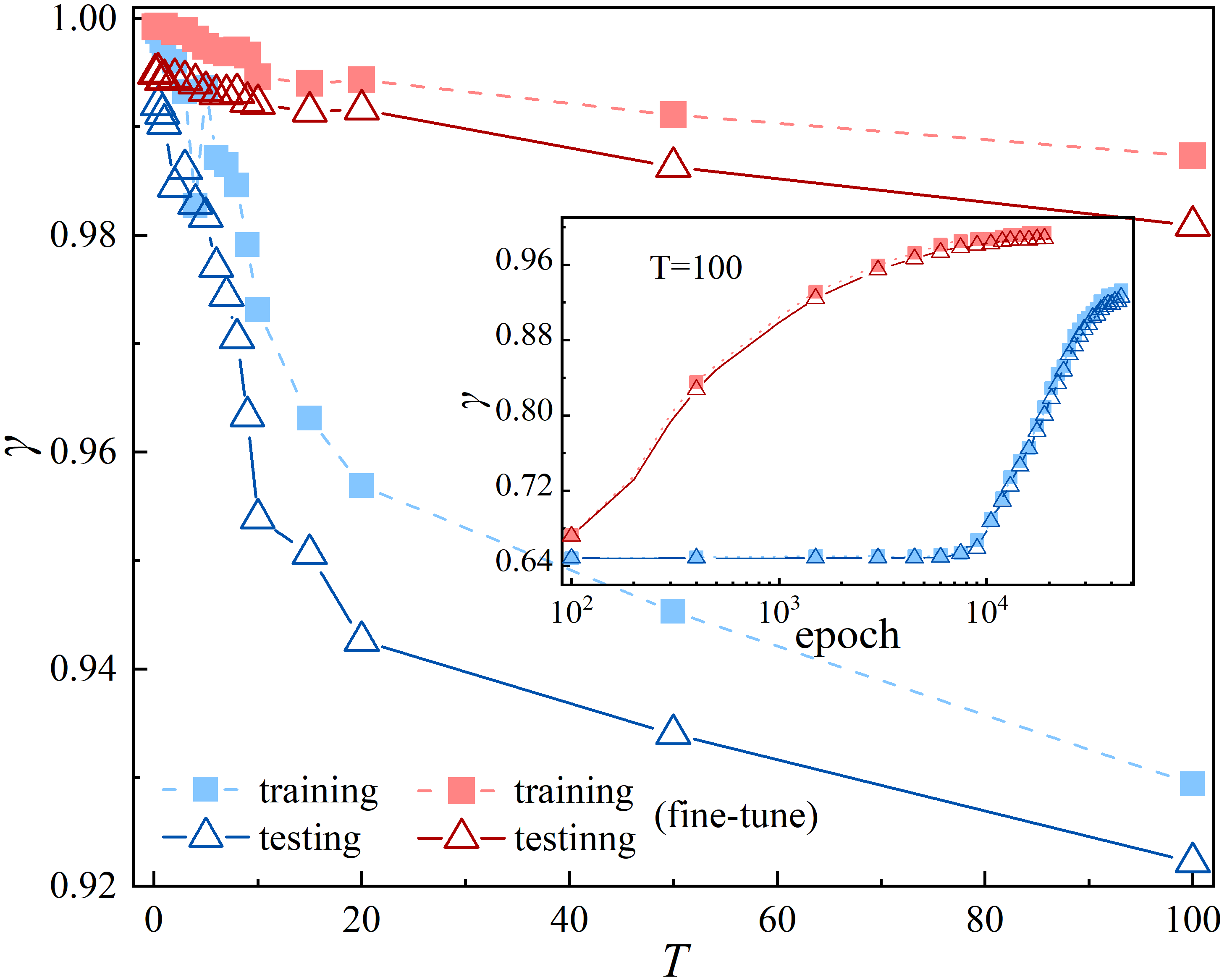}
	\caption{(Color online) The training and testing accuracies ($\gamma$) versus the temperature ($T$). The red symbols show the accuracies by fine-tuning on the CNN trained by the low-temperature data, and the blue ones show the accuracies obtained by training a randomly initialized CNN (without fine-tuning). The inset shows the accuracy for different numbers of epochs with and without fine-tuning at $T=100$. The ``barren plateau'' without fine-tuning lasts to about $10^{4}$ epochs, and the convergent accuracy is much lower than that by using fine-tuning.}
	\label{fig-3}
\end{figure}

As the temperature increases, the thermal fluctuations grow and suppress the correlations along both space and time. Different lattices will exhibit almost identical dynamics when the thermal fluctuations dominate, making accurate estimations unlikely by, e.g., the statistic methods. 

Fig. \ref{fig-3} shows how the accuracy $\gamma$ changes with temperature $T$. We fix the number of training and testing instances to be $N_{\text{train}}=10000$ and $N_{\text{test}}=3000$, respectively. By independently training the CNN at each temperature, the accuracy $\gamma$ drops fast for about $T>5$ (blue symbols). This is somewhat expected since the thermal fluctuations suppress the relevance between the lattice topology and the magnetic momenta. 

By fine-tuning the CNN trained by the low-temperature data, accurate predictions can be made even at the very high temperatures (say $T=100$). Specifically, we pre-train the CNN with the instances at $T=0.1$, and then with those at the target temperatures. Note the pre-training temperature and the details in the fine-tuning strategy are flexible.

The key is to let the CNN  pre-learn some knowledge on the relevance between the topology and magnetic data when the relevance is strong (say at a low temperature). If we directly use the high-temperature data to train a randomly initialized CNN, the gradients of the variational parameters in the CNN would nearly vanish. This leads to a ``barren plateau'' induced by the strong thermal fluctuations, akin to the ``quantum barren plateau'' caused by noises~\cite{Wang2021}. It lasts to about $10^{4}$ epochs as shown in the inset of Fig.~\ref{fig-3}. Even the optimization manages to get out of the plateau after more than $10^{4}$ epochs, it still converges to an accuracy that is much lower than that using the fine-tuning. 

To further test the generalization (i.e., the ability of processing the unlearnt data), we introduce the generalization set as the instances evolved from different lattices. It means that the CNN learnt from many instances evolved with the lattices in the training set, but none from the lattices in the generalization set. Thus, the generalization set disobeys the \textit{i. i. d.} and is unbalanced to certain extent. 

\begin{figure}[tbp]
	\centering
	\includegraphics[angle=0,width=1\linewidth]{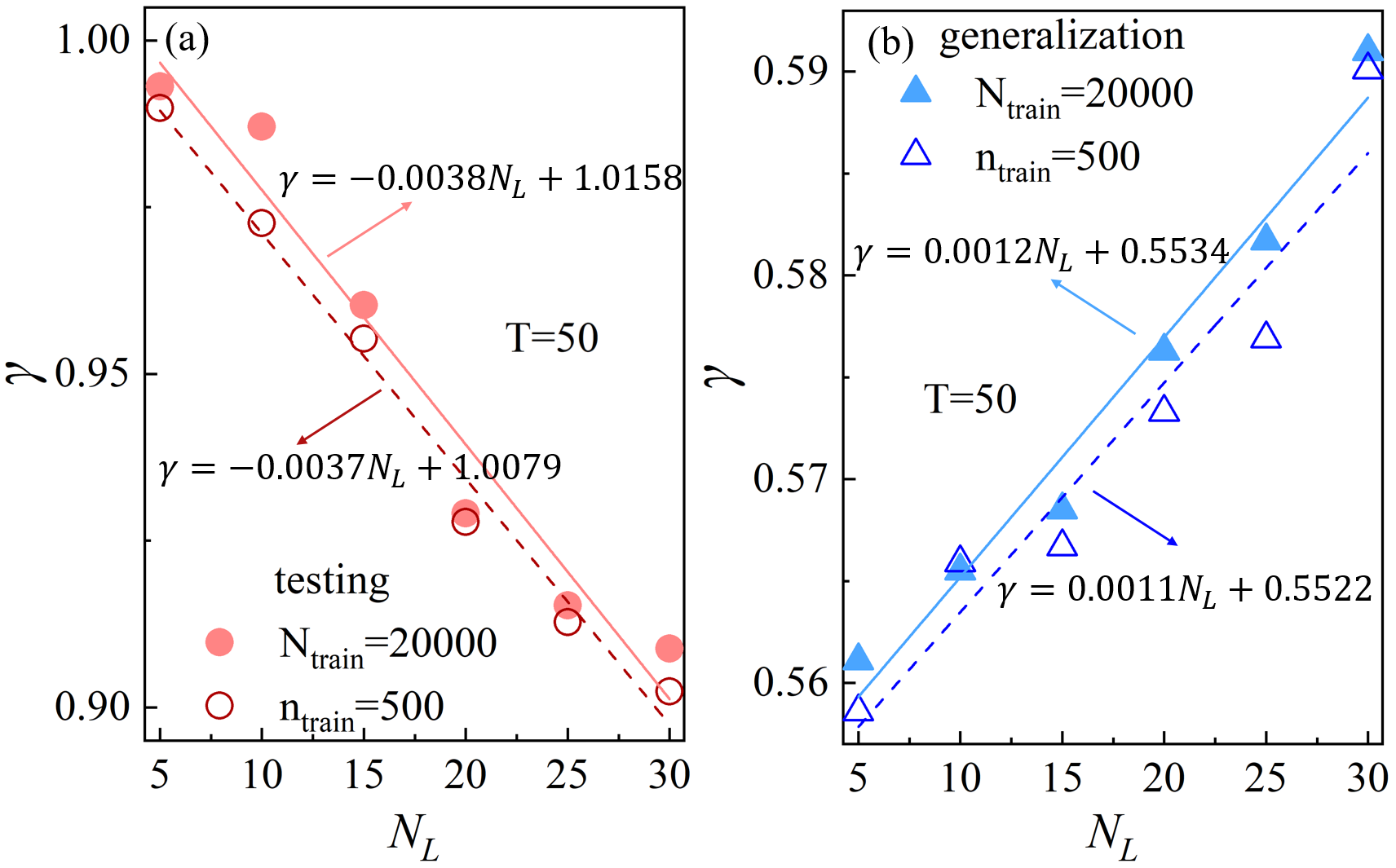}
	\caption{(Color online) The (a) testing and (b) generalization accuracies ($\gamma$) versus the number of lattices in the training set ($N_{L}$) at a high temperature $T=50$. The solid symbols show the results by fixing the total number of training instances to be $N_{\text{train}}=20000$. The fitting errors for the testing and generalization sets are $R^2=0.9557$ and $0.9729$, respectively. The hollow symbols show those where each lattice generates $n_{\text{train}}=500$ training instances, with $R^2=0.9859$ and $0.9279$, respectively.}
	\label{fig-5}
\end{figure}

Fig.~\ref{fig-5} shows the testing and generalization accuracies versus the number of lattices $N_{L}$ in the training set at a high temperature $T=50$. We fix the total number of the training instances with $N_{\text{train}} = 20000$, or fix the training number of instances for each lattice as $n_{\text{train}} = 500$. For the latter, we have $N_{\text{train}} = n_{\text{train}} N_{L}$. In both cases, the testing accuracy decreases linearly with $N_{L}$ as 
 \begin{eqnarray}
	\gamma =  aN_{L} + b,
	\label{eq-linear}
\end{eqnarray}
with $a=-0.0038$ and $-0.0037$, $b=1.0158$ and $1.0079$, respectively [Fig.~\ref{fig-5} (a)]. Such decreases of testing accuracy are typically due to the higher requirement on the learnability of the CNN to capture the information from more lattices. The testing accuracy can be improved by increasing the complexity (e.g., depth or number of variational parameters) of the CNN. 

In contrast, the generalization accuracy is much lower that the testing accuracy, but increases linearly with $a=0.0012$ and $0.0011$, $b=0.5534$ and $0.5522$, respectively [Fig.~\ref{fig-5} (b)]. Be aware that $N_{L}=30$ is still a small number compared with the total number of possible lattices with $L=12$ spins. The data are definitely unbalanced with the \textit{i.i.d.} disobeyed, which causes strong effects on the accuracy for the generalization set. It would require a significant number of instances to satisfy the \textit{i.i.d.} since the numbers of both the possible lattices and the initial states grow (nearly) exponentially with the system size. Such a double-exponential degrees of freedom puts forward a challenge on the applications of ML to statistic systems. 

Our data indicate that the CNN have still learnt the information on dealing with the unlearnt lattices, as the predictions are obviously more accurate than the random guesses and the accuracy is growing with $N_{L}$. The learnt information is not about the topology, but has to be about the ``Glauber dynamics'' itself. Since CNN and in general the deep ML are non-interpretable, interpreting and further enhancing the learnt information about the intrinsic dynamics remain to be open questions.

\textit{Summary.---} In this work, we propose the deep ML of the dynamic data of kinetic Ising model for predicting the lattice topology. The trained deep convolutional neural network (CNN) maps the time-dependent magnetic momenta to the adjacent matrix, differing from the existing statistic methods that require more data to estimate, e.g., correlations and transfer entropy. Our deep learning scheme achieves high accuracy even with strong thermal fluctuations at high temperatures, where the perturbative and statistic methods in general fail. Specifically, the fine-tuning on the CNN trained with the low-temperature data help to avoid the ``barren plateau'' caused by the dominative fluctuations. The generalization of CNN is investigated. The CNN exhibits high accuracy on the testing set where the principle of \textit{i.i.d} is obeyed. The challenge emerges when we consider to predict on the unlearnt lattice topology. This is essentially due to the ``double-exponential'' degrees of freedom from both the possible lattices and initial states, which makes the satisfaction of \textit{i.d.d} extremely expensive. Our method can be flexibly applied to other statistic systems and network models. The CNN can be replaced by other ML models such as the graph/recursive neural networks ~\cite{scarselli_graph_2009, elman_finding_1990} and the more interpretable models such as tensor networks (e.g., Refs.~[\onlinecite{SS16TNML, HWFWZ17MPSML, LRWP+17MLTN}]). 

\textit{Acknowledgment.} This work was supported by NSFC (Grant No. 12004266, No. 11834014, and No. 11774300), and Foundation of Beijing Education Committees (No. KM202010028013). SJR acknowledges the support from the Academy for Multidisciplinary Studies, Capital Normal University.

%

\end{document}